%% file: main.tex
\documentclass[letterpaper, 10 pt, conference]{ieeeconf}  

\IEEEoverridecommandlockouts                              

\usepackage{graphics} 
\usepackage{epsfig} 
\usepackage{times} 
\usepackage{amsmath} 
\usepackage{amssymb}  

\title{\LARGE \bf
Adaptive and Explainable Deployment of Navigation Skills\\via Hierarchical Deep Reinforcement Learning
}

\author{Kyowoon Lee$^{*1}$, Seongun Kim$^{*2}$ and Jaesik Choi$^{2}$
\thanks{*Equal contribution}
\thanks{$^{1}$ Department of Computer Science and Engineering, Ulsan National Institute of Science and Technology, Ulsan, Republic of Korea. {\tt\small leekwoon@unist.ac.kr}.}%
\thanks{$^{2}$ Graduate School of Artificial Intelligence, Korea Advanced Institute of Science and Technology, 291 Daehak-ro, Daejeon 34141, Republic of Korea. {\tt\small \{seongun, jaesik.choi\}@kaist.ac.kr} Correspondence to: Jaesik Choi}
}

\begin{document}

\maketitle
\thispagestyle{empty}
\pagestyle{empty}

\begin{abstract}

For robotic vehicles to navigate robustly and safely in unseen environments, it is crucial to decide the most suitable navigation policy. However, most existing deep reinforcement learning based navigation policies are trained with a hand-engineered curriculum and reward function which are difficult to be deployed in a wide range of real-world scenarios. In this paper, we propose a framework to learn a family of low-level navigation policies and a high-level policy for deploying them. The main idea is that, instead of learning a single navigation policy with a fixed reward function, we simultaneously learn a family of policies that exhibit different behaviors with a wide range of reward functions. We then train the high-level policy which adaptively deploys the most suitable navigation skill. We evaluate our approach in simulation and the real world and demonstrate that our method can learn diverse navigation skills and adaptively deploy them. We also illustrate that our proposed hierarchical learning framework presents explainability by providing semantics for the behavior of an autonomous agent.


\end{abstract}


\input{1_introduction}

\input{2_related_work}
\input{3_problem_formulation}
\input{4_approach}
\input{5_experiments}
\input{6_conclusion}

\input{0_acknowledgment}

\bibliographystyle{IEEEtran}
\bibliography{references}

\addtolength{\textheight}{-12cm}   

\end{document}

%% file: 1_introduction.tex
\section{INTRODUCTION}




Autonomous navigation of mobile robots has gained much interest due to a wide variety of important applications in industry. These include assistive robots \cite{feil2011socially}, a last-mile delivery \cite{hoffmann2018regulatory}, a guidance at airports \cite{kayukawa2019bbeep}, and warehouse navigation \cite{xiao2017robot}, among others. To navigate a robot to a desired goal in complex, cluttered, and highly dynamic environments with multiple static and dynamic obstacles, a reliable and robust navigation policy is essential.

Traditional approaches to address the navigation problem typically consist of a series of modules, each of which is specifically designed for solving a particular sub-task of navigation problems such as human detection, prediction of future trajectories of humans, and path planning \cite{gupta2018social, liu2020robot}.
However, these approaches require extensive computational efforts and rely on manually engineered parameters, which limit the ability of mobile robots to operate in previously unseen environments, or across different robotic platforms, as the modular system suffers from a lack of generalization and sub-optimal performance \cite{sathyamoorthy2020frozone}. In this regard, deep reinforcement learning (DRL) approaches have recently been proposed as learning based solutions for autonomous navigation, which directly map raw sensor observations to controls and have shown remarkable results compared to traditional methods and robustness to sensor noise \cite{long2018towards, fan2019getting, jin2020mapless, dugas2021navrep}. However, existing DRL-based approaches require hand-engineered reward functions that can be exceedingly time consuming. Moreover, in challenging scenarios, DRL-based approaches with a fixed reward function easily get stuck in local optima which makes complex situations including navigating across narrow corridors or dense crowds and turning corners, unsolvable.

\begin{figure}
    \centering
    \includegraphics[width=0.99\linewidth]{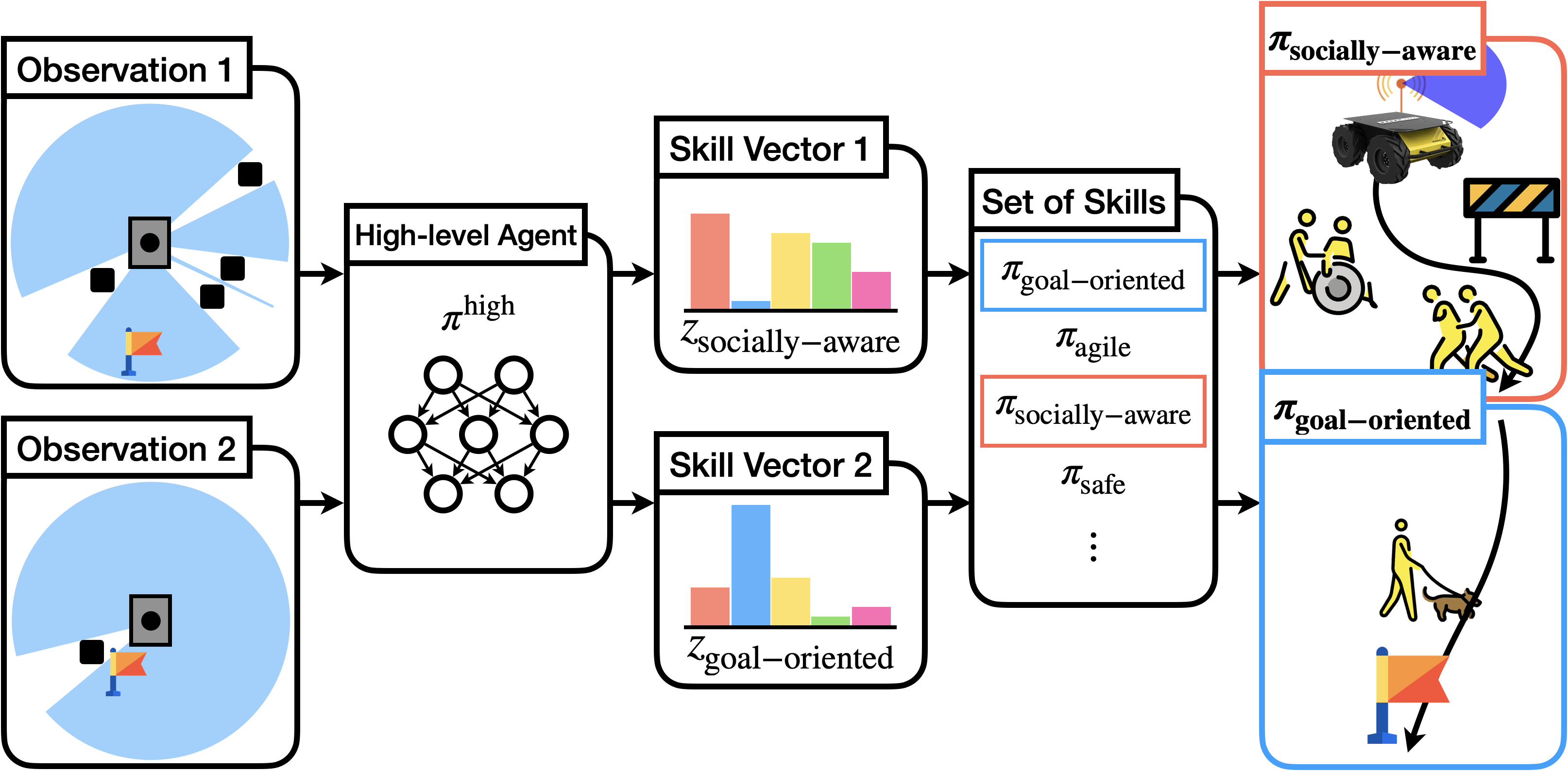}
    \caption{An overview of the proposed hierarchical framework of navigation skills. A high-level policy invokes low-level navigation skills from a raw sensory observation. A low-level policy is adopted from a continuous skill vector among the infinite number of skills, which then drives a robot while understanding the context.}
    \label{fig:main_figure}
\end{figure}

Another potential problem of existing DRL-based robot navigation is that a policy represented by a deep neural network often lacks transparency and cannot provide explanations on decision-making reasons. Recent works investigate DRL methods which are interpretable and provide decision-making reasons \cite{lee2018modular, trivedi2021learning}. However, they are only applicable in video games or in simple simulation environments. Another line of research provides explanations in real-world robots by highlighting input features that a deep network most refers to when making a decision \cite{kim2021explaining}. Nevertheless, this approach has the limitation that it is not directly relevant to increasing the performance of the policy model.


To address the aforementioned problems, we develop a framework of learning to adaptively deploy navigation skills that are explainable with hierarchical reinforcement learning (HRL).\footnote{Source codes are available at https://github.com/leekwoon/hrl-nav} Specifically, our approach can be divided into two phases. The first phase trains a family of low-level navigation policies, each of which is optimized for a particular skill vector in a continuous representation. This skill vector is associated with a corresponding reward function. For example, a skill vector that imposes an additional penalty for the robot to be more cautious with obstacles can exhibit socially-aware navigation behavior, while a skill vector encouraging the robot getting closer to its goal may result in more aggressive behavior. In the second phase, we train the high-level policy which deploys the most suitable navigation skill for every time step. The main contributions of this work can be summarized as follows:

\begin{itemize}
    \item Proposal of a hierarchical reinforcement learning approach which learns diverse navigation skills and deploys them.
    \item Extensive evaluation of our approach on various scenarios including a real-world robot navigation which demonstrates effectiveness and explainability of our approach.
\end{itemize}


%% file: 2_related_work.tex
\section{RELATED WORK}

\subsection{Reinforcement Learning for Navigation}

Recent advances in deep reinforcement learning have enabled the solving of complex navigation tasks from raw sensory measurements. Long et al. \cite{long2018towards} propose a DRL-based multi-agent navigation framework to train collision avoidance policy with proximal policy optimization (PPO) \cite{schulman2017proximal} using 2D LiDAR observation. This DRL-based collision avoidance policy is further integrated into a hybrid control framework \cite{fan2018fully} and a conventional global planner from robot operating system (ROS) \cite{guldenring2020learning}. A similar approach is proposed in \cite{jin2020mapless}, however, additional ego-safety and social-safety rewards are used to consider human-awareness. Apart from using 2D LiDAR observation, navigation that learns from RGB images has also been studied \cite{zhu2017target, gupta2017cognitive}.

\subsection{Reinforcement Learning with Sparse Rewards}

Existing approaches which handle sparse rewards in reinforcement learning (RL) can be divided into two categories, curriculum reinforcement learning and reward shaping. In the context of curriculum reinforcement learning, the implicit curriculum uses goal relabeling techniques which randomly sample goals from failed trajectories \cite{andrychowicz2017hindsight, fang2018dher}. The explicit curriculum considers the difficulty of goals during training, e.g., by generating the goal further from the start in navigation or increasing the number of obstacles \cite{florensa2018automatic, long2018towards}. The reward shaping technique modifies the reward signal by learning a parameterized dense reward function \cite{ng1999policy}. Sibling rivalry \cite{trott2019keeping} uses a model-free, dynamic reward shaping method that preserves optimal policies on sparse-rewards tasks. On the other hand, AutoRL \cite{chiang2019learning} uses large-scale hyperparameter optimization to shape the reward.

\subsection{Hybrid Control framework for Navigation}

To further improve the robustness and effectiveness of navigation policy by leveraging the strength of multiple local planners, a hybrid control framework has been widely adopted which uses a high-level switching controller to manage a set of low-level control rules. For example, Jin et al. \cite{jin2017stable} propose hand-designed switching rules to combine goal-navigation and obstacle avoidance, Shucker et al. \cite{shucker2007switching} introduce switching rules to handle challenging cases where collisions or noises cannot be handled appropriately, and Fan et al. \cite{fan2018fully} classify the scenarios into three cases by considering a robot's sensor measurement and using both a PID and DRL-based controller. Most similar to our work, Kastner et al. \cite{kastner2022all} propose a hierarchical navigation system integrating model-based and learning-based local planners by training an agent which decides between multiple local planners. However, the approach uses only one DRL-based local planner with the traditional model-based planners. As a result, navigation systems following this approach are often strongly dependent on deployed model-based planners. Instead, we train a family of policies that exhibit different behaviors and focus on how to adaptively deploy them.

%% file: 3_problem_formulation.tex
\section{PROBLEM FORMULATION}

HRL decomposes a general RL into a hierarchy of multiple sub-problems which themselves are RL problems \cite{hengst2010hierarchical}. Higher-level problems abstract an original RL problem and adopt which lower-level problems to solve, while lower-level problems are defined to solve the original problem given an abstraction from the higher-level problems.

In this work, we decompose a problem of learning navigation skills into a hierarchy of two sub-problems, a high-level problem and a low-level problem. The high-level problem is to find a macro policy which understands the context of a navigation environment and adopts which navigation skills to invoke. The macro policy observes a current state and demonstrates the context that the agent encounters. The low-level problem is to discover a set of navigation skills that drives a mobile robot to reach a target position. Each low-level navigation skill issues a command by giving primitive actions such as motor velocities, conditioned on the context distinguished by the macro policy.

We formalize each sub-problem as a Markov decision process (MDP), in particular, a goal-conditioned MDP \cite{schaul2015universal} for learning general-purpose navigation skills.
A high-/low-level problem is modeled as an MDP $\langle \mathcal{S}^\mathrm{high/low}, \mathcal{A}^\mathrm{high/low}, \mathcal{G}, P^\mathrm{high/low}, R_g^\mathrm{high/low}, \gamma \rangle$, where $\mathcal{S}$ is the set of states, $\mathcal{G}$ is the set of goals, $\mathcal{A}$ is the set of actions, $P: \mathcal{S} \times \mathcal{A} \times \mathcal{S} \to [0, +\infty)$ is the transition probability, $R_g: \mathcal{S} \times \mathcal{A} \times \mathcal{A} \to \mathbb{R}$ is the goal-conditioned reward function, $\gamma \in [0, 1]$ is the discount factor, and each superscript $\mathrm{high}$ and $\mathrm{low}$ represents a notation for the high-level and the low-level problem respectively.

\begin{figure*}
    \centering
    \includegraphics[width=0.8\linewidth]{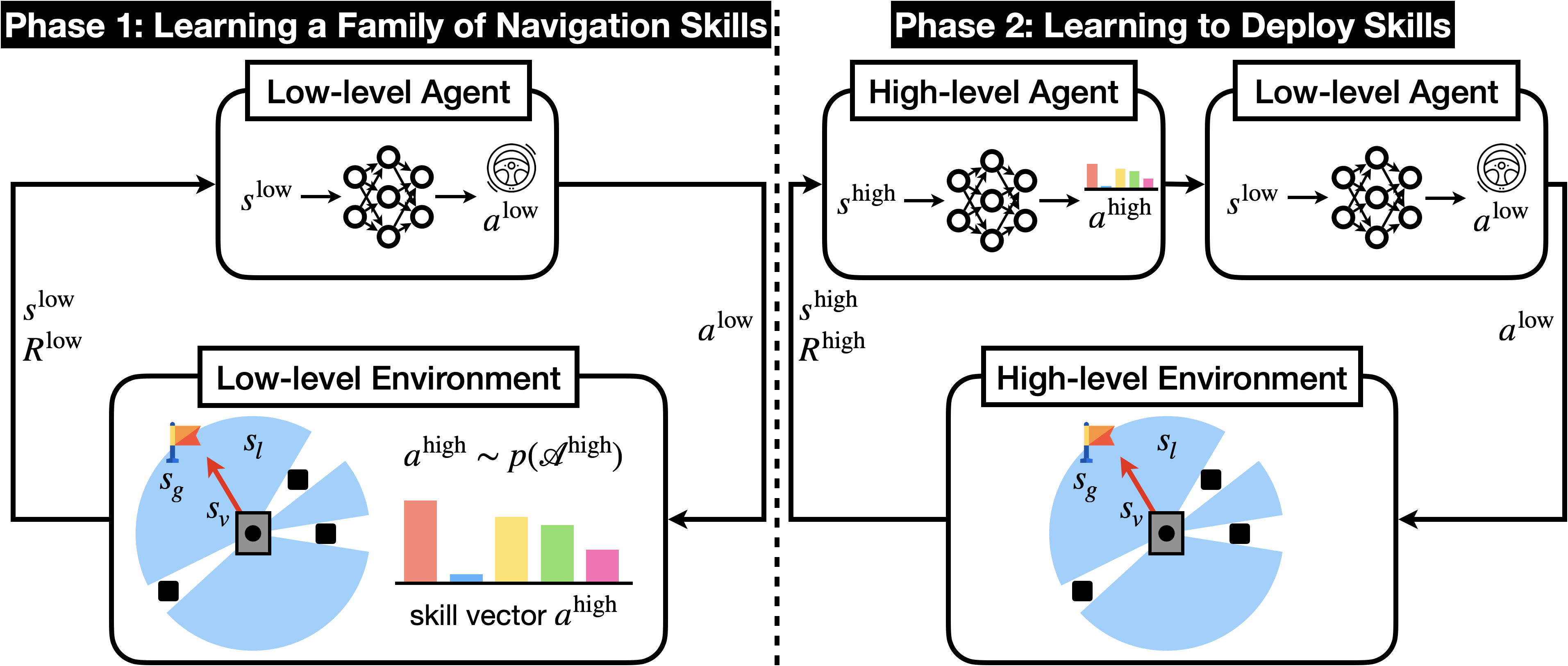}
    \caption{A two-phase training procedure of the proposed hierarchical framework. A family of navigation skills is trained in the first phase, while a high-level policy is trained in the second phase. To train the infinite number of navigation skills in the first phase, a single policy network which is conditioned on a continuous skill vector is utilized. The skill vector $a^\mathrm{high}$ is sampled from a fixed distribution per episode to construct $s^\mathrm{low}$ along with a scan observation.
    The high-level agent is then trained by taking actions from the low-level skill deployed by the high-level agent.}
    \label{fig:procedure}
\end{figure*}

\begin{figure}
    \centering
    \includegraphics[width=0.85\linewidth]{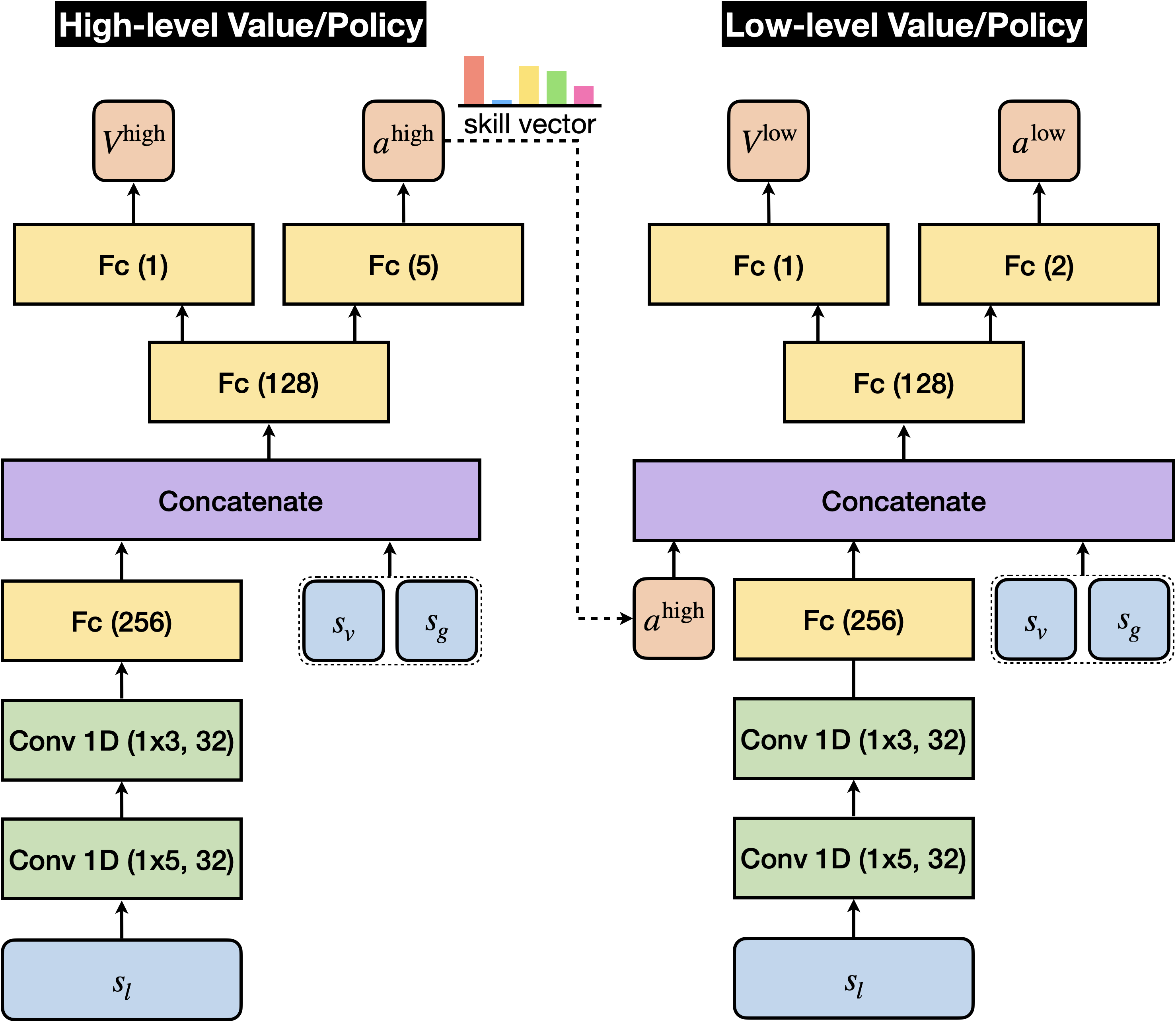}
    \caption{Actor-critic architecture used for learning a family of navigation skills (left) and learning to deploy skills (right). $\mathrm{Conv}$ denotes a convolutional layer and $\mathrm{Fc}$ denotes a fully-connected layer.}
    \label{fig:network}
\end{figure}

To build the hierarchy, an action from the high-level agent $a^\mathrm{high} \in \mathcal{A}^\mathrm{high}$ is observed as a state by the low-level agent, where this high-level action implies the context of the environment. In addition to $a^\mathrm{high}$, the low-level agent observes the same state as the high-level agent, which results in $s^\mathrm{low} = (s^\mathrm{high}, a^\mathrm{high}) \in \mathcal{S}^\mathrm{low}$. While interacting with the environment and observing the state $s^\mathrm{low}$, each low-level agent receives a reward from a reward function $R_g^\mathrm{low}(s^\mathrm{low}, a^\mathrm{low}, a^\mathrm{high})$ that consists of multiple reward terms which we will describe in more detail in the following section. To effectively train each low-level agent to present discriminative primitive actions conditioned on $a^\mathrm{high}$, they are trained with different reward functions by weighting each reward term in $R_g^\mathrm{low}$ by $a^\mathrm{high}$.

Under this problem definition, the optimal low-level policies and the high-level policy in the hierarchy are achieved by an off-the-shelf RL algorithm which maximizes a corresponding value function formulated as follows:
\begin{align}
    &V^{\mathrm{low}}(s^\mathrm{low}, g | \pi^\mathrm{high})
    \triangleq \mathbb{E}_{\substack{a_t^\mathrm{low} \sim \pi^\mathrm{low}(a_t^\mathrm{low}|s_t^\mathrm{low}, g),\\a_t^\mathrm{high} \sim \pi^\mathrm{high}(a_t^\mathrm{high}|s_t^\mathrm{high}, g),\\s_{t+1}^\mathrm{low} \sim P(s_{t+1}^\mathrm{low}|s_{t}^\mathrm{low}, a_{t}^\mathrm{low})}} \nonumber \\
    &\quad \Bigg[\sum_{t=0}^{\infty} \gamma^t R_g^\mathrm{low}(s_t^\mathrm{low}, a_t^\mathrm{low}, a_t^\mathrm{high}) \Big| s_0^\mathrm{low} = s^\mathrm{low}  \Bigg],
\end{align}
\begin{align}
    &V^{\mathrm{high}}(s^\mathrm{high}, g | \pi^\mathrm{low})
    \triangleq \mathbb{E}_{\substack{a_t^\mathrm{high} \sim \pi^\mathrm{high}(a_t^\mathrm{high}|s_t^\mathrm{high}, g),\\a_t^\mathrm{low} \sim \pi^\mathrm{low}(a_t^\mathrm{low}|s_t^\mathrm{low}, g),\\s_{t+1}^\mathrm{high} \sim P(s_{t+1}^\mathrm{high}|s_{t}^\mathrm{high}, a_{t}^\mathrm{high})}} \nonumber \nonumber \\
    &\quad \Bigg[\sum_{t=0}^{\infty} \gamma^t R_g^\mathrm{high}(s_t^\mathrm{high}, a_t^\mathrm{high}, a_t^\mathrm{low}) \Big| s_0^\mathrm{high} = s^\mathrm{high}  \Bigg].
\end{align}

%% file: 4_approach.tex
\section{APPROACH}


In the hierarchy, the high-level agent and the low-level agent interact with each other. To train agents in the hierarchy, we decompose a training procedure into two phases. In this section, we describe how each component of MDP is defined and how the agent is trained for each training phase.


\subsection{Learning a Family of Navigation Skills}

In the first phase, we learn a family of low-level navigation policies, each optimized for a particular skill vector in continuous representation. We present a detailed setup of phase 1 as follows:

\subsubsection{State space} A state $s_t^\mathrm{low}$ at time step $t$ consists of four components: the raw 2D LiDAR measurements $s_t^l \in \mathbb{R}^{512}$ casting 512 rays over 360$^{\circ}$ with up to 6m range, the linear and angular velocity of the robot $s_t^v \in \mathbb{R}^2$, the relative goal state $s_{t}^g \in \mathbb{R}^2$ represented in the polar coordinate, and the high-level action $a_t^\mathrm{high} \in \mathbb{R}^5$. We call $a_t^\mathrm{high}$ \textit{a skill vector} since it induces a specific behavior by weighting a number of reward terms.

\subsubsection{Action space} Considering nonholonomic kinematics constraints, an action $a_t^\mathrm{low}  \in \mathbb{R}^2$ is composed of the linear velocity $v_t \in [0, 0.5]$ and angular velocity $w_t \in [-0.64, 0.64]$ which is normalized into $[-1, 1]$ in the neural network.

\subsubsection{Reward function} Our goal is to learn a family of navigation policies with a diverse set of behavioral characteristics. Thus, we set the reward function parameterized by the skill vector $a^\mathrm{high}$ as follows: 
\begin{align*}
\label{eq:reward}
    &R_t^\mathrm{low}(s_t^\mathrm{low},a_t^\mathrm{low},a_t^\mathrm{high}) \\
    &= r_{\mathrm{success}}+a^{\mathrm{high}}[r_{\mathrm{collision}}\;r_{\mathrm{progress}}\;r_{v}\;r_{w}\;r_{\mathrm{safety}}]^\top.
\end{align*}
The first term, $r_{\mathrm{success}}$, is the success reward, which is 1 if the goal is achieved and 0 otherwise. The collision reward, $r_{\mathrm{collision}}$ is -1 when the robot collides with obstacles and 0 otherwise. The progress reward, $r_{\mathrm{progress}}$ is the difference between the previous distance to the goal and the current distance to the goal. The driving reward, $r_{v}$ is the positive linear speed which encourages the robot to drive as fast as possible. The turning reward, $r_{w}$ is the negative angular speed for smooth driving. Finally, inspired by the social-safety zone in \cite{jin2020mapless}, $r_{\mathrm{safety}}$ is the safety reward that gives a penalty with value $1 - \frac{d_t}{r + 0.5}$ when the robot enters the safety zone, where $d_t$ is the distance to the closest obstacle and $r$ is the robot radius. 

\subsubsection{Network architecture} The neural network architecture used for learning a family of navigation skills is illustrated in the right of Fig. \ref{fig:network}. We follow the architecture in the prior work \cite{long2018towards}, except that the policy and the value function are additionally parameterized by the skill vector $a^\mathrm{high}$ which is concatenated with the output of the third layer, the current velocity $s_v$, and the relative goal state $s_g$.

\begin{figure}
    \centering
    \includegraphics[width=0.99\linewidth]{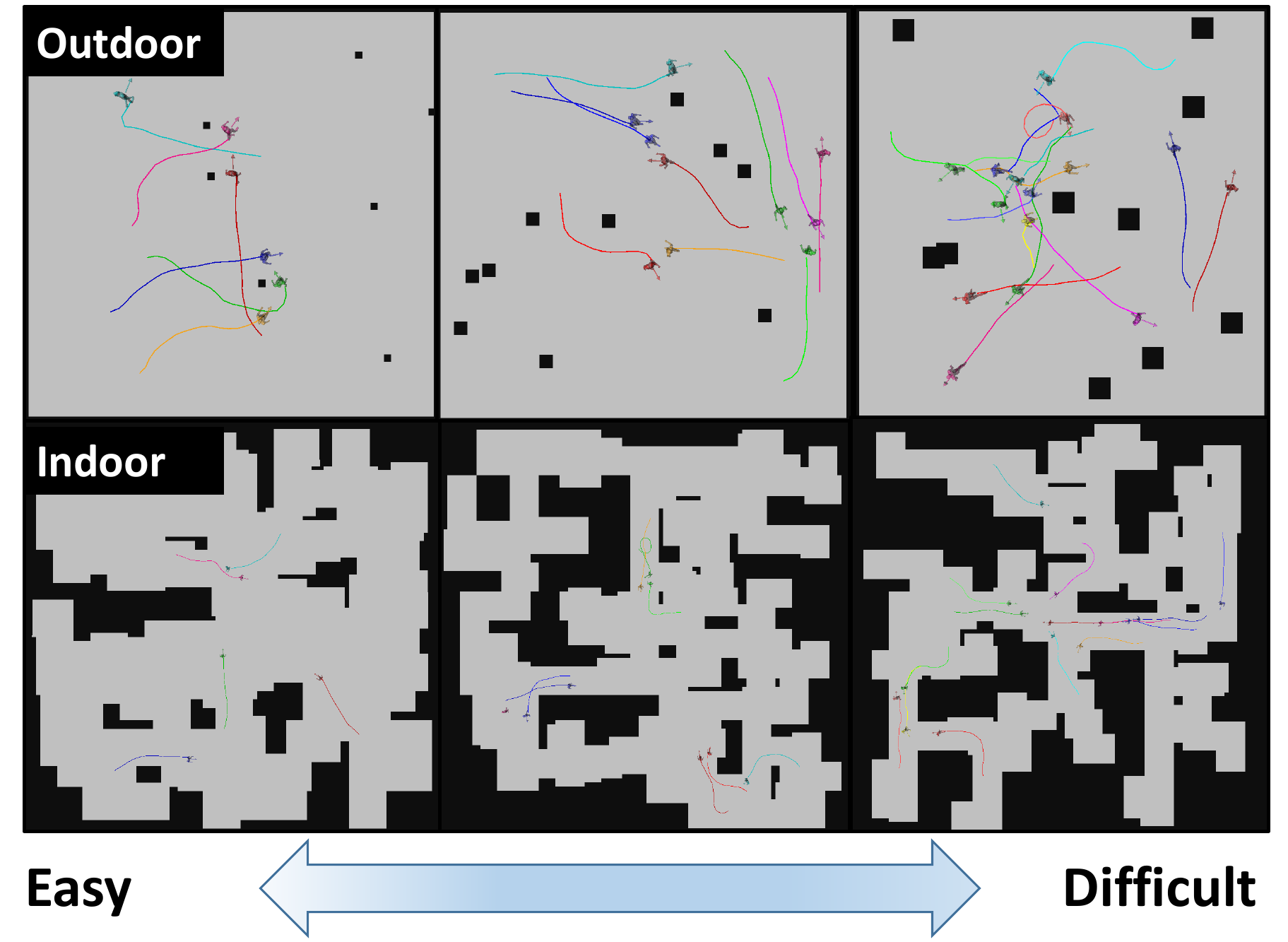}
    \caption{Training environments used to train the policies, with both randomized maps and obstacles. We randomize training environments so that the agent learns diverse skills. To control the difficulty, we manipulate the number of humans, obstacles, and corridor and the width of corridor.}
    \vspace{-0.15in}
    \label{fig:training_env}
\end{figure}



\subsubsection{Training procedure}
As each individual policy is trained under the similar structure of reward function, we use a single deep neural network to train the entire family of low-level policies simultaneously. Inspired by a universal policy \cite{yu2017preparing}, we train a policy $\pi^\mathrm{low}:(s_t^\mathrm{low}, a_t^\mathrm{high}) \rightarrow a_t^\mathrm{low}$ that is conditioned on both state of the robot $s_t^\mathrm{low}$ and the skill vector $a_t^\mathrm{high}$. In contrast to existing approaches which learn a single policy, at the beginning of the episode, we sample a skill vector from predefined distribution and fix it during the rollout to train corresponding policy $\pi_{a_t^\mathrm{high}}: s_t^\mathrm{low} \rightarrow a_t^\mathrm{low}$, which we call a skill.

\subsection{Learning to Deploy Skills}

\begin{figure}
    \centering
    \includegraphics[width=0.95\linewidth]{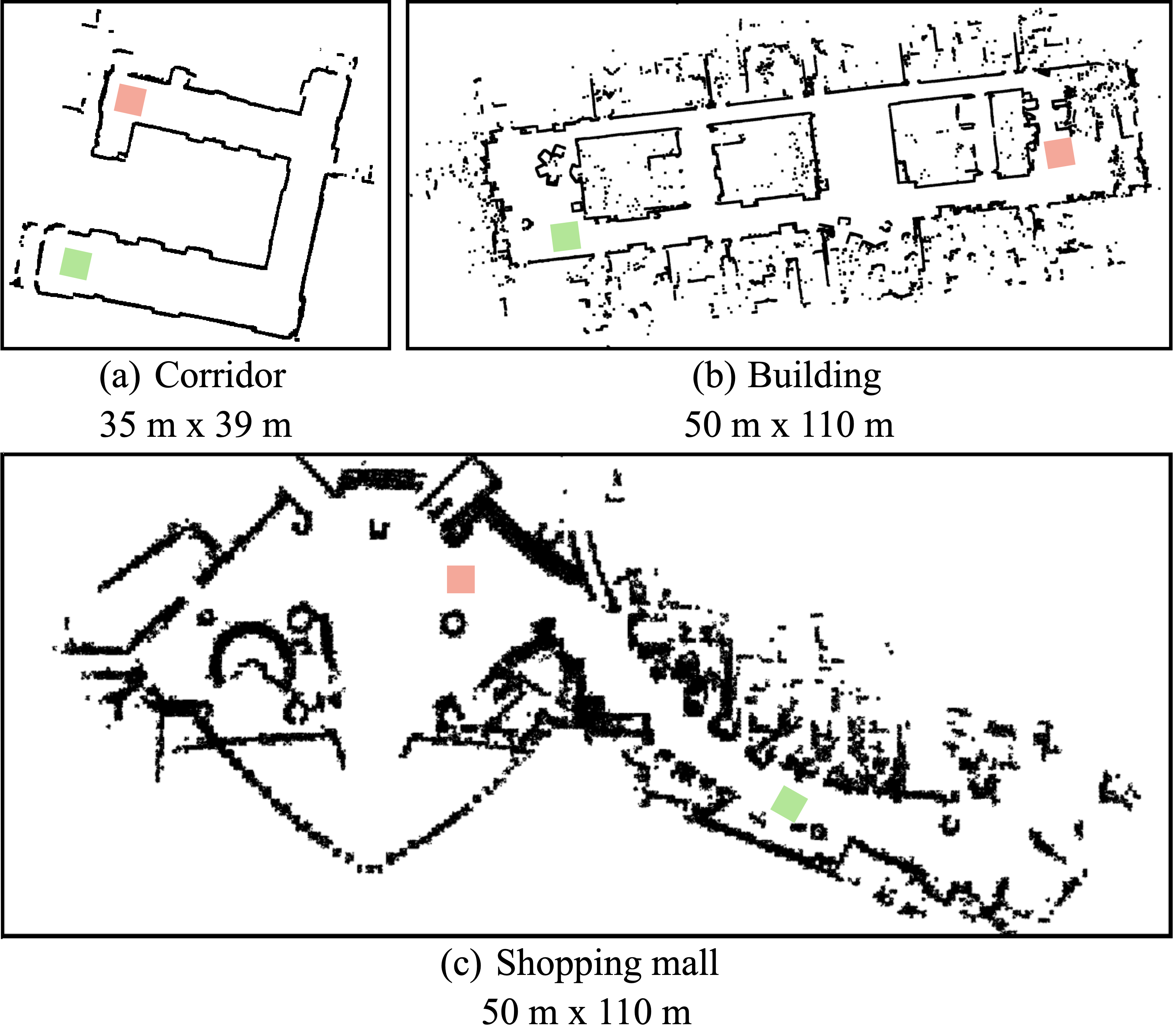}
    \caption{Unseen environments used to evaluate the policies. The robot navigates from the green square to the goal is located on the red square.}
    \label{fig:evaluation_env}
    \vspace{-0.15in}
\end{figure}

\begin{figure*}
    \centering
    \includegraphics[width=0.95\linewidth]{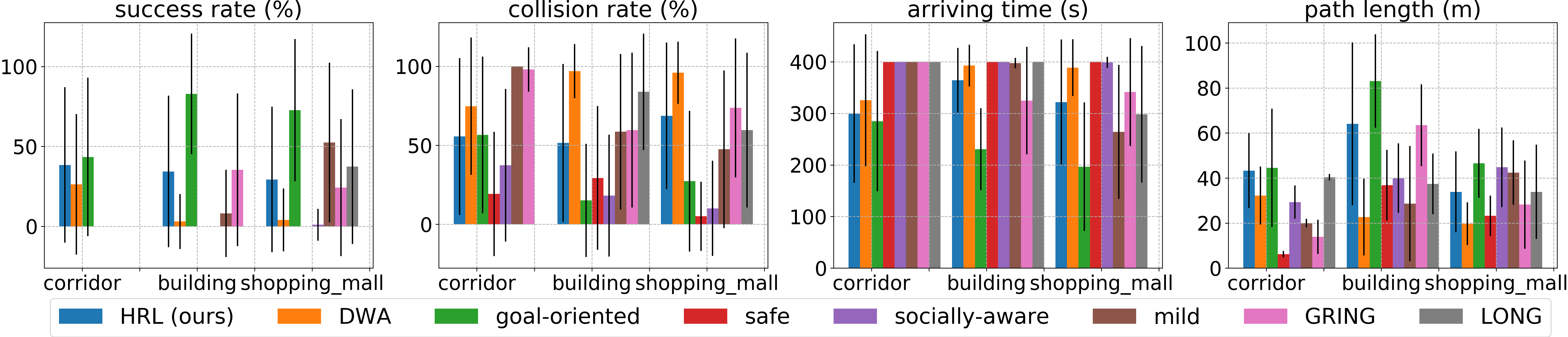}
    \caption{Quantitative results in three unseen environments with 100 episodes. We evaluate the proposed method by comparing four metrics: success rate, collision rate, time to reach the goal, and path length. Our hierarchical method (blue bar) presents comparative performance in all evaluation environments.}
    \label{fig:quantitative_1}
\end{figure*}

The second phase of our approach is to learn a high-level policy to adaptively deploy the navigation skills learned in phase 1. The following provides a detailed formulation.

\subsubsection{State space} A state $s_t^\mathrm{high}$ composed of three components,  $s_t^l$, $s_t^v$ and $s_t^g$, which are used in the first phase.

\subsubsection{Action space} An action $a_t^\mathrm{high} \in \mathbb{R}^5$ is a skill vector that decides the learned low-level behavior characteristic.


\subsubsection{Reward function} The objective of the second phase is to minimize the time to reach the desired goal and we use the following sparse reward function:
\begin{align*}
& R_t^\mathrm{high}(s_t^\mathrm{high},a_t^\mathrm{high},a_t^\mathrm{low}) = \begin{cases} 0 &  \text{if reach the goal } \\ -1 &  \text{otherwise},\end{cases}
\end{align*}
where the agent gets the reward of 0 when the goal is achieved, and -1 otherwise.

\subsubsection{Network architecture} We parameterize each of the policies and value functions with the architecture from the prior work \cite{long2018towards}, except that the policy predicts a particular skill vector that will be used by the low-level policy. We further feed the output of high-level policy as an input to the low-level policy to infer the command velocity of the robot during a rollout as shown in Fig. \ref{fig:network}.

%

\subsubsection{Training procedure}
After finishing training the set of navigation skills, we can further perform hierarchical control by training the high-level policy. During a rollout, the high-level policy predicts the skill vector which decides the behavior characteristic. This skill vector is observed as an additional input to the low-level policy which outputs the command velocity of a robot. Note that, in the second phase, gradients flow only through the high-level policy, not the low-level policy. Due to a problem with the sparse reward setting where a positive reward is only provided when the goal is achieved, we use the hindsight experience replay (HER) technique \cite{andrychowicz2017hindsight} which provides an effective way to handle the sparse reward challenge by revisiting previous states in the replay buffer and storing additional trajectories with hindsight goals for generating reward signals.


%

%% file: 5_experiments.tex
\section{EXPERIMENTS AND RESULTS}

\begin{figure}
    \centering
    \includegraphics[width=0.9\linewidth]{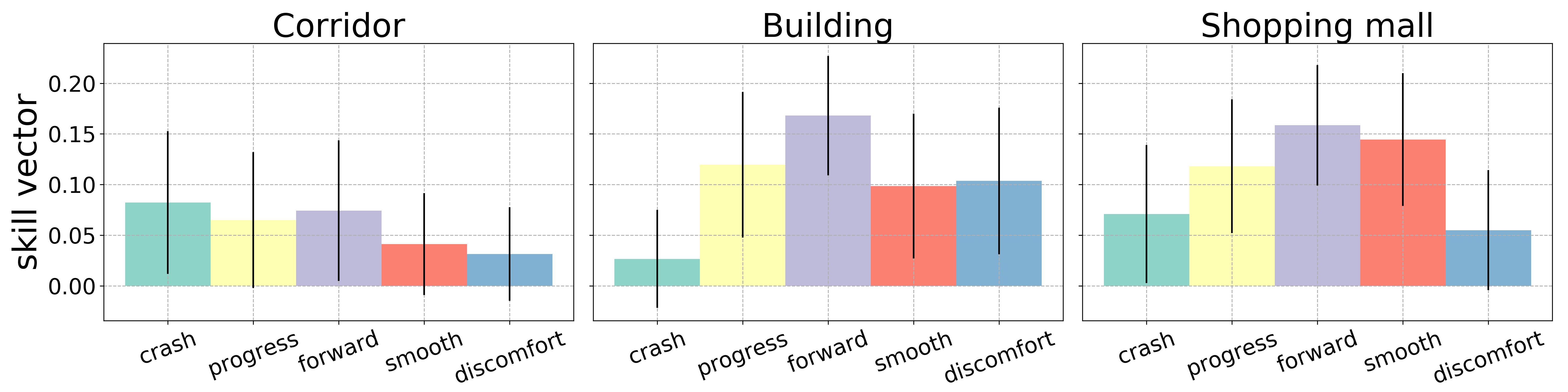}
    \caption{Results of the qualitative analysis. We compare the average value of the skill vector with standard deviation among three evaluation environments. The high-level agent deploys slow but careful skills in the first environment, while it deploys smooth but fast skills in the other environments.}
    \label{fig:qualitative_1}
    \vspace{-0.1in}
\end{figure}



\subsection{Setup}
We train a policy using an OpenAI-gym-compatible simulator that we specially design to integrate it into the robot operating system (ROS) with the goal of open-sourcing it for easy comparison of various approaches including state-of-the-art learning-based approaches and conventional ones (see Fig. \ref{fig:training_env}). To create a scenario with various difficulty levels, we adopt the map generation method from \cite{kastner2022all} to generate indoor (50m$\times$50m) and outdoor (20m$\times$20m) maps. We randomize
the noise on sensors, and static and dynamic obstacles where we control the motion of the dynamic obstacles using a learning-based collision avoidance method that extends the prior work \cite{long2018towards} with $A^{*}$ global planner to make it follow the sub-goal generated from the global path for more realistic long-range behavior. We terminate the simulation when reaching the goal, colliding with an obstacle, or exceeding 1000 time steps.

\begin{figure}
    \centering
    \includegraphics[width=0.95\linewidth]{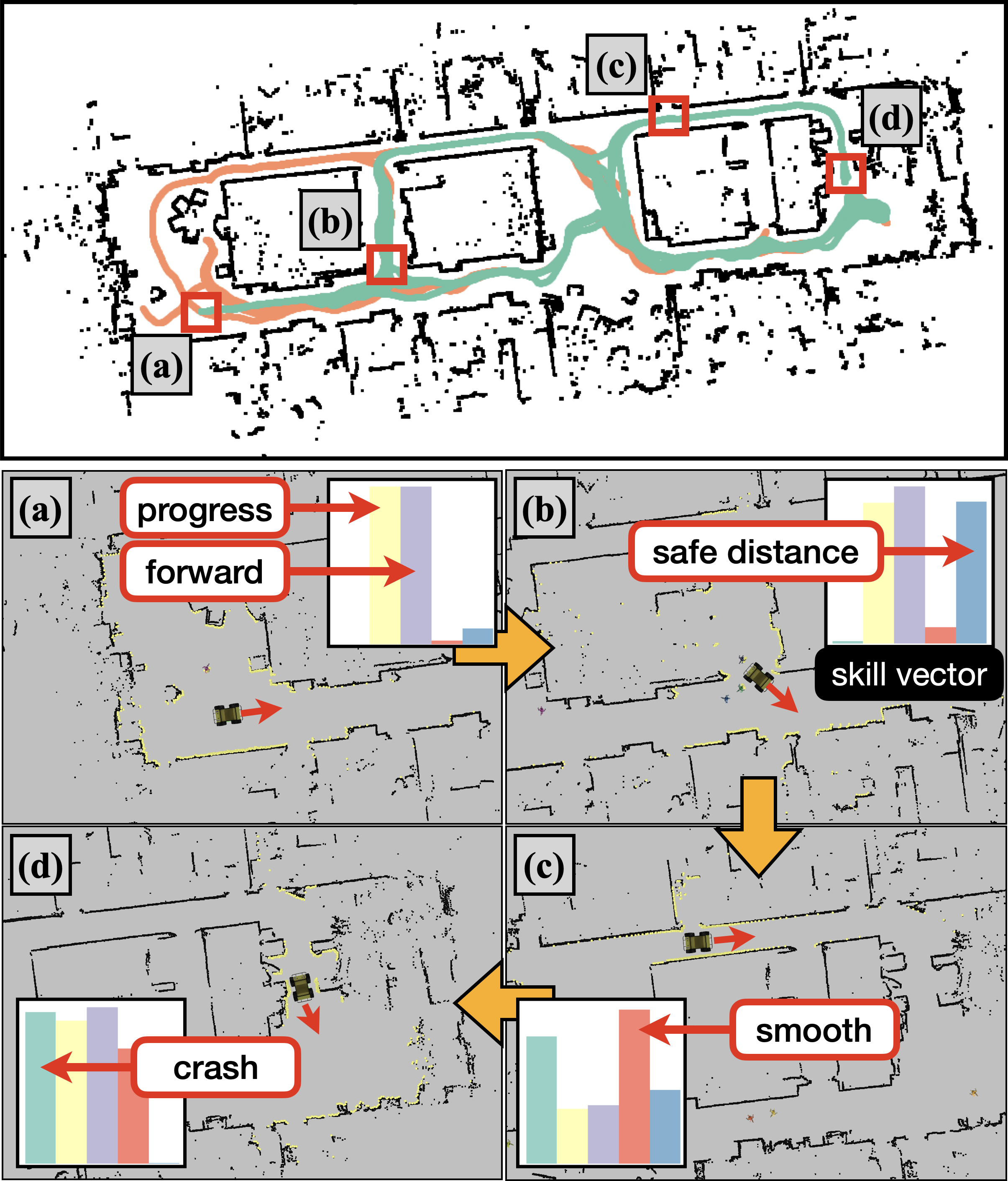}
    \caption{Trajectories of the our approach in the unseen building environment (top) and skill vectors in particular situations (bottom).}
    \label{fig:qualitative_building}
    \vspace{-0.15in}
\end{figure}

\begin{figure*}
    \centering
    \includegraphics[width=0.9\linewidth]{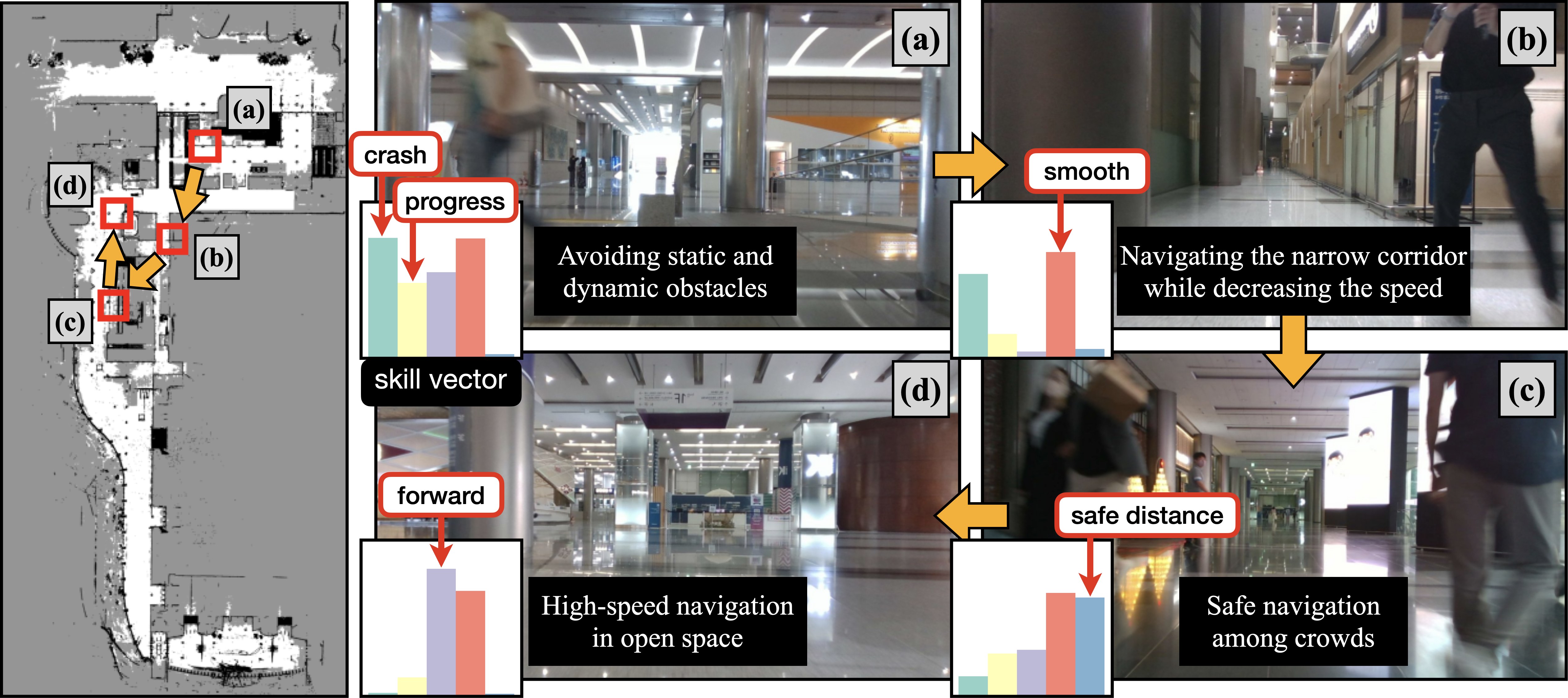}
    \caption{Demonstration of our approach in unseen real scenarios. The scenarios are navigating in (a) cluttered environment, (b) narrow corridor, (c) dense crowd and (d) open space. The skill vectors are predicted by transferred high-level policy.}
    \label{fig:real}
    \vspace{-0.15in}
\end{figure*}

To train the policies, we use soft actor-critic (SAC) \cite{haarnoja2018soft} with an automatically tuned entropy coefficient, employing the Adam optimizer \cite{kingma2014adam} with a learning rate 1e-4 for both the actor and the critic, a target smoothing coefficient of 5e-4, and a replay buffer size of 1M.

We compare our approach with a traditional model-based approach dynamic window approach (DWA), two DRL-based approaches LONG \cite{long2018towards} and GRING \cite{guldenring2020learning}, and four agents trained with a fixed skill vector which we term $\pi_\mathrm{goal-oriented}$, $\pi_\mathrm{socially-aware}$, $\pi_\mathrm{safe}$, and $\pi_\mathrm{mild}$ where each policy is trained with the skill vector of $[0, 0.2, 0.1, 0, 0]$, $[0.1, 0, 0, 0.1, 0.2]$, $[0.2, 0, 0, 0.1, 0.1]$, and $[0.1, 0.1, 0.1, 0.2, 0]$ respectively.



\subsection{Quantitative Analysis}
We evaluate the trained policies in three unseen large environments (see Fig. \ref{fig:evaluation_env}). These scenarios include (a) a corridor with 4 dynamic obstacles, (b) a building with 9 dynamic obstacles, and (c) a shopping mall with 8 dynamic obstacles which are designed to evaluate the ability of the agent including global planning, collision avoidance, and safe navigation, among other real-world situations. We validate the effectiveness of our method with four metrics: success rate, collision rate, arriving time, and path length.
Fig. \ref{fig:quantitative_1} shows the evaluation results with 100 episodes in each scenario. To solely evaluate the performance of learned navigation policy, all approaches do not use a global planner that is frequently used to handle a long-range navigation challenge.
A traditional model-based approach DWA fails in dense crowds scenarios. Further, in all evaluation environments, our approach shows comparative performance to other baselines including DRL-based agents and agents trained with the fixed skill vector, except for goal-oriented policy. LONG and GRING present comparative results in densely crowded environments where a local navigation policy can solve but performance drops significantly for a situation where a global navigation policy is required.
An interesting observation is that $\pi_\mathrm{goal-oriented}$ performs well in long-range and dynamic environments without crash or safety rewards typically used in prior works. We assume that this is due to the goal-oriented reward, which encourages exploration and goal achievement.
It would be an appealing research direction to study the role of reward terms rigorously.

\subsection{Qualitative Analysis}

To qualitatively analyze the effectiveness of the proposed approach, we interpret the average skill vector along the trajectory for each unseen evaluation environment in Fig. \ref{fig:qualitative_1}. In the corridor environment, there is a long wall. To successfully reach the goal, the robot needs to detour a long range. To this end, the high-level agent deploys skills that present slow but careful motor commands with relatively high crash skill value.
On the other hand, in the other two environments, the building and the shopping mall, there are little static obstacles on the straight line from the initial position of the robot to the target position. In these environments, the skill vector has large progress and forward values so that the robot presents agile movement. In addition, the high-level agent demonstrates smooth motor command through the smooth skill value.
The average discomfort skill value in the last two crowded environments is relatively large, which validates the effectiveness of our approach.

The whole trajectories of our approach in the building environment are shown in Fig. \ref{fig:qualitative_building} where green lines represent trajectories of successes. The skill vectors demonstrate the interpretable deployment of learned navigation skills. At the beginning where there are few obstacles around the robot as shown in (a) of Fig. \ref{fig:qualitative_building}, progress and forward skill values are huge for fast navigation towards the goal. In the middle of the navigation shown in (b), though progress and forward skill values are still large, the discomfort value increases when the robot is surrounded by humans, which would lead the robot to avoid collision with them. When the robot enters the narrow corridor in (c), the smooth value increases to prevent oscillatory behaviors. Around the goal which is close to obstacles in (d), crash is high to safely arrive the goal. The skill vector presents the same tendency across the other evaluation scenarios. We refer the reader to the videos in supplementary material for further information.



\subsection{Real-World Experiment}

To demonstrate the validity of the proposed approach, we collect the expert trajectory with our robot which has two 270$^\circ$ LiDARs that are merged together to cover 360$^\circ$. We then evaluate how well the high-level policy understands the context in a real-world shopping mall. Fig. \ref{fig:real} shows the skill vector generated by high-level policy for different scenarios. The high-level policy encourages the robot to keep a safe distance from crowds and to decrease the speed in narrow corridors. On the other hand, the goal-oriented navigation skill with high speed is preferred in open space.


%% file: 6_conclusion.tex
\section{CONCLUSION}

In this paper, we propose a hierarchical learning framework to train reinforcement learning based navigation policies that can present explainability by providing semantics of a behavior of an autonomous agent and can reduce the effort to design hand-engineered reward functions. The evaluation results on unseen environments including real-world cases demonstrate the adaptive and explainable deployment of learning navigation skills.

%% file: 0_acknowledgment.tex
\section*{ACKNOWLEDGMENT}

This work was supported by the Industry Core Technology Development Project, 20005062, Development of Artificial Intelligence Robot Autonomous Navigation Technology for Agile Movement in Crowded Space, funded by the Ministry of Trade, Industry \& Energy (MOTIE, Republic of Korea) and was supported by Institute of Information \& communications Technology Planning \& Evaluation (IITP) grant funded by the Korea government (MSIT) (No. 2022-0-00984, Development of Artificial Intelligence Technology for Personalized Plug-and-Play Explanation and Verification of Explanation, No.2019-0-00075, Artificial Intelligence Graduate School Program (KAIST)).
